\documentclass[conference]{IEEEtran}
\IEEEoverridecommandlockouts
\usepackage{cite}
\usepackage{amsmath,amssymb,amsfonts}
\usepackage{algorithmic}
\usepackage{graphicx}
\usepackage{textcomp}
\usepackage{xcolor}
\usepackage{hyperref}
\usepackage{multirow}
\usepackage{makecell}
\usepackage{arydshln}
\usepackage{booktabs}
\def\BibTeX{{\rm B\kern-.05em{\sc i\kern-.025em b}\kern-.08em
    T\kern-.1667em\lower.7ex\hbox{E}\kern-.125emX}}
\begin{document}

\title{Text-driven Online Action Detection}

\author{
Manuel Benavent-Lledo$^{a,*}$, David Mulero-Perez$^{a}$, David Ortiz-Perez$^{a}$, Jose Garcia-Rodriguez$^{a,b,c}$\\
$^{a}$Department of Computer Technology, University of Alicante, Spain\\
$^{b}$ValgrAI - Valencian Graduate School and Research Network of Artificial Intelligence, Valencia, Spain\\
$^{c}$Institute of Informatics Research, University of Alicante, Alicante, Spain \\
$^*$Corresponding author: {\tt mbenavent@dtic.ua.es} \\}

\maketitle

\begin{abstract}
Detecting actions as they occur is essential for applications like video surveillance, autonomous driving, and human-robot interaction. Known as online action detection, this task requires classifying actions in streaming videos, handling background noise, and coping with incomplete actions. Transformer architectures are the current state-of-the-art, yet the potential of recent advancements in computer vision, particularly vision-language models (VLMs), remains largely untapped for this problem, partly due to high computational costs. In this paper, we introduce TOAD: a Text-driven Online Action Detection architecture that supports zero-shot and few-shot learning. TOAD leverages CLIP (Contrastive Language-Image Pretraining) textual embeddings, enabling efficient use of VLMs without significant computational overhead. Our model achieves 82.46\% mAP on the THUMOS14 dataset, outperforming existing methods, and sets new baselines for zero-shot and few-shot performance on the THUMOS14 and TVSeries datasets.
\end{abstract}


\section{Introduction} \label{sec:intro}
Detecting the occurrence of an action as soon as it happens is a critical task with a wide range of real-world applications, including: video surveillance~\cite{oad-surveillance,GARCIARODRIGUEZ20114413}, autonomous driving~\cite{Kim_2019_CVPR,ramanishka2018CVPR} or human-robot interaction~\cite{oad-robot,GOMEZDONOSO2017105}, among others. This task, known as online action detection, has attracted much interest in recent years.

\begin{figure}
    \centering
    \includegraphics[width=\linewidth]{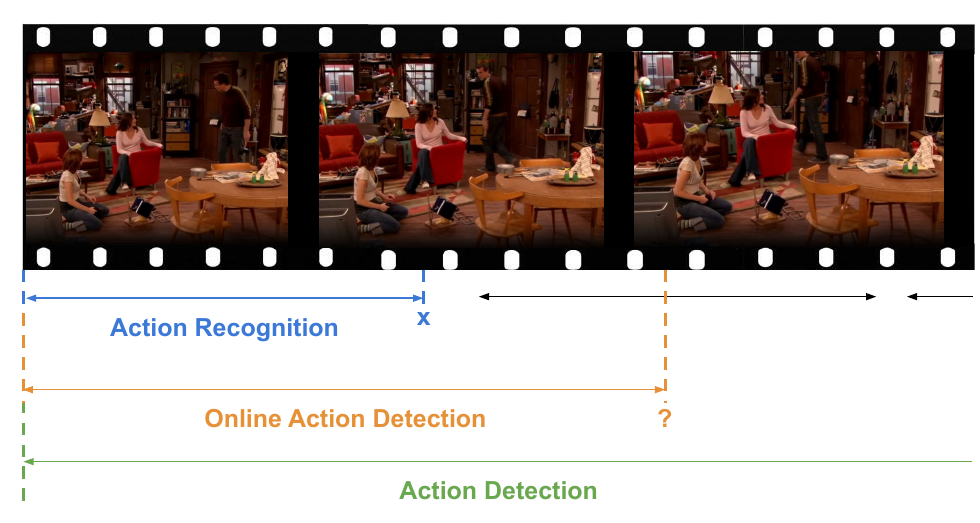}
    \caption{Comparison of action recognition (blue), online action detection (orange), and action detection (green). Black segments represent the annotated actions. Note that online action detection may involve multiple actions, background frames without annotated actions, and/or incomplete actions. In contrast, action recognition focuses on classifying predefined annotated segments. And action detection aims to localize and classify actions in the video after it has been fully viewed. Video frames extracted from \cite{de2016online}.}
    \label{fig:oad}
\end{figure}

The online action detection problem presents some unique challenges that are not encountered in other action understanding tasks (\emph{e.g.}, action recognition, or action detection). The goal is to classify every frame in a streaming video. As represented in Figure~\ref{fig:oad}, unlike action recognition, online action detection contends with a wide variety of negative data, consisting of background frames such as unannotated segments and scene changes. Additionally, unlike conventional action detection, which processes the entire untrimmed video, the online setting must handle partial or incomplete actions due to the absence of future information.

This task, first introduced by De Geest et al.~\cite{de2016online}, requires modeling long-range temporal dependencies to interpret recent events. For years, recurrent neural networks (RNNs) have been the dominant approach in this area~\cite{Eun_2020_CVPR,gao2017red,Xu_2019_ICCV,An_2023_ICCV}. However, despite their theoretical ability to retain past information, RNNs are known to often struggle to effectively capture long-range temporal dependencies.

In contrast, the self-attention mechanism of transformer architectures has set a new standard in natural language processing~\cite{zhao2023survey}, image analysis~\cite{dosovitskiy2021image}, and video analysis~\cite{arnab2021vivit,piergiovanni2022rethinking}. This mechanism excels at modeling long-range temporal dependencies, making transformer based approaches the state-of-the-art for online action detection as well~\cite{Wang_2021_ICCV,10.1145/3582649.3582656,xu2021long,zhao2022testra}. However, these methods often incur significant computational cost due to the self-attention mechanism, hindering their adoption across different datasets.

To address these challenges, vision-language models (VLMs) offer a promising alternative by emulating human learning. These models associate semantic information from visual appearances with language, rather than relying on a predefined set of classes. VLMs have demonstrated state-of-the-art results in a variety of video understanding tasks~\cite{wang2021actionclip,bike,radford2021learning,cheng2023vindlu,ju2022prompting,papalampidi2023simple,li2023strong,wu2023revisiting}. By enabling zero-shot and few-shot approaches, this type of architecture is able to reduce the need for large and densely labeled datasets for training downstream tasks.

VLMs were first explored in~\cite{10.1007/978-3-031-61137-7_6} for the online action detection problem. Based on those findings, in this work we present TOAD, the first Text-driven model for Online Action Detection. As remarked in~\cite{10.1007/978-3-031-61137-7_6,wu2023revisiting} contrastive learning favors large batch sizes resulting in a higher computational cost. Instead, using textual embeddings to initialize a classifier can also provide state-of-the-art results with a significant reduction in computational cost~\cite{wu2023revisiting}. Our open-source implementation can be found on GitHub\footnote{\url{https://github.com/3dperceptionlab/TOAD}}.

In summary, the contributions of this paper are the following:

\begin{itemize} 
\item We present TOAD, a novel text-driven online action detection architecture that achieves state-of-the-art performance on the THUMOS14 dataset using only RGB data. 
\item Leveraging the capabilities of VLMs for data-efficient learning, our approach establishes a new baseline for zero-shot and few-shot online action detection, and offers a robust foundation for future research. 
\item Our comprehensive ablation study validates both the proposed method and its individual components. \end{itemize}

The remainder of the paper is organized as follows: Section~\ref{sec:relwork} reviews relevant literature on vision-language models and online action detection. Section~\ref{sec:method} details the components of TOAD. Section~\ref{sec:exp} presents the experimental setup, results, and ablation study to demonstrate the efficacy of our method. Finally, Section~\ref{sec:concl} offers the conclusions drawn from this work.
\section{Related Works} \label{sec:relwork}
This section provides an overview of pertinent literature on online action detection and the integration of language models for video understanding.

\subsection{Vision-Language Models}
The capabilities of large language models (LLMs) in temporal modeling and feature representation have been extensively explored~\cite{brown2020language, touvron2023llama, devlin2018bert}. These models have been applied to various vision tasks, most notably for image and video captioning. While traditional methods focus on generating textual descriptions of videos~\cite{8627985}, recent research has explored the use of language models to improve action recognition. Contrastive learning approaches inspired by CLIP (Contrastive Language-Image Pretraining)~\cite{radford2021learning}—such as ActionCLIP~\cite{wang2021actionclip} and VideoCLIP~\cite{xu2021videoclip}—aim to align visual and textual embeddings for more integrated representations.

VideoCLIP follows CLIP’s pre-training paradigm, adding temporal information to enable several downstream tasks. On the other side, ActionCLIP proposes to address the downstream task of action recognition, as the upstream task by using the proposed ``pre-train, prompt, fine-tune'' paradigm. The features from a CLIP pre-trained model are temporally aggregated and aligned with textual features of action classes, enabling zero-shot and few-shot action recognition.

Nonetheless, the computational requirements of such architectures have prompted a shift towards more efficient methods such as~\cite{cheng2023vindlu,ju2022prompting,papalampidi2023simple,li2023strong,wu2023revisiting}. Wu et al.~\cite{wu2023revisiting} highlight the necessity of huge batch sizes in contrastive learning. Instead, they propose to initialize a frozen classifier for action recognition using CLIP's textual embedding. Alternatively,  BIKE~\cite{bike} presents an innovative framework designed to enable bi-directional cross-model knowledge transfer between vision and language models, aiming to enhance action recognition in videos.

\subsection{Online Action Detection}
The objective of online action detection is to identify an action the moment it occurs, meaning each frame of a video stream is classified in real-time without any knowledge of future frames. Initially introduced by De Geest et al.~\cite{de2016online}, previous similar tasks comprised early event detection and frame-level frame detection as remarked in \cite{Hu2022}.

Addressing this problem requires modeling long-range temporal dependencies. For years, recurrent neural networks have been the leading approach in this area. \cite{Eun_2020_CVPR,gao2017red,Xu_2019_ICCV,An_2023_ICCV}. However, as previously introduced they tend to struggle with long-range temporal dependencies, and have been outperformed by transformer based architectures which exploit the self-attention mechanism.

Building on this foundation, OadTR~\cite{Wang_2021_ICCV} was the first transformer introduced for online action detection. It uses pre-extracted spatial features from both RGB and optical flow frames. The proposed method captures temporal relationships by aggregating the early fused RGB and optical flow embeddings through a transformer encoder. Simultaneously, a transformer decoder leverages the encoder's output to predict upcoming frames, thereby anticipating future actions. This approach enhances the model's performance by incorporating auxiliary information from the predicted future representations, leading to more accurate results. Similarly, the authors of LightTR~\cite{10.1145/3582649.3582656} proposed an encoder-decoder transformer architecture with simplified self-attention layers for more efficient online action detection.

Colar~\cite{yang2022colar} proposes to model category-level features as complimentary guidance to the temporal modeling. To do so, a dynamic branch models temporal dependencies while a static one compares frames to category exemplars, \emph{i.e.} frames representing the same action to the category particularities. The authors propose an effective exemplar-consultation mechanism that first measures the similarity between a frame and exemplary frames, and then aggregates exemplary features based on the similarity weights.

Xu et al.~\cite{xu2021long} present LSTR, a long-short term transformer for online action detection. A long- and short-term memory mechanism is employed to model prolonged sequences of data comprising up to 8 minutes (long) or 8 seconds (short). The LSTR encoder compresses the long-term memory which is used as input alongside the short-term memory for the LSTR decoder which outputs the probability vectors. The memory management algorithm follows a FIFO logic for storing the representations.

TesTRa~\cite{zhao2022testra} tackled the challenge of computational complexity, which typically grows quadratically with the length of the considered temporal dynamics. To overcome this complexity issue, the authors enhanced the mechanism proposed in LSTR by incorporating two types of temporal smoothing kernels: a box kernel and a Laplace kernel. These kernels allow the model to reuse most of the computations from frame to frame, significantly reducing the processing required for each new frame. This approach ensures a constant update time between frames, enabling TesTRa to run faster than equivalent sliding-window approaches.

Alternatively, to improve the performance of recurrent neural networks, the authors of MiniROAD~\cite{An_2023_ICCV} proposed a novel approach by assigning non-uniform weights to the loss computed at each time step. This strategy was developed after an investigation on why RNNs struggle to capture long-range dependencies. By emphasizing certain time steps more than others, the model can better mimic the conditions of the inference stage during training, leading to improved efficiency and achieving equal or better results compared to traditional methods.

WOAD~\cite{Gao_2021_CVPR} introduces a weakly supervised framework for online action detection that can be trained using only video-level labels. This framework employs two jointly trained modules: a temporal suggestion generator and an online action recognizer. LSTMs are used to aggregate information, allowing the model to effectively learn and detect actions without the need for frame-level annotations.

Leveraging the previously introduced ``pre-train, prompt, fine-tune'' paradigm for vision-language models, \cite{10.1007/978-3-031-61137-7_6} is the first approach for zero-shot and few-shot online action detection. Despite the results falling short of state-of-the-art performance, it is presented as a promising alternative in data-limited scenarios.

\section{TOAD Architecture} \label{sec:method}
\begin{figure*}
    \centering
    \includegraphics[width=\textwidth]{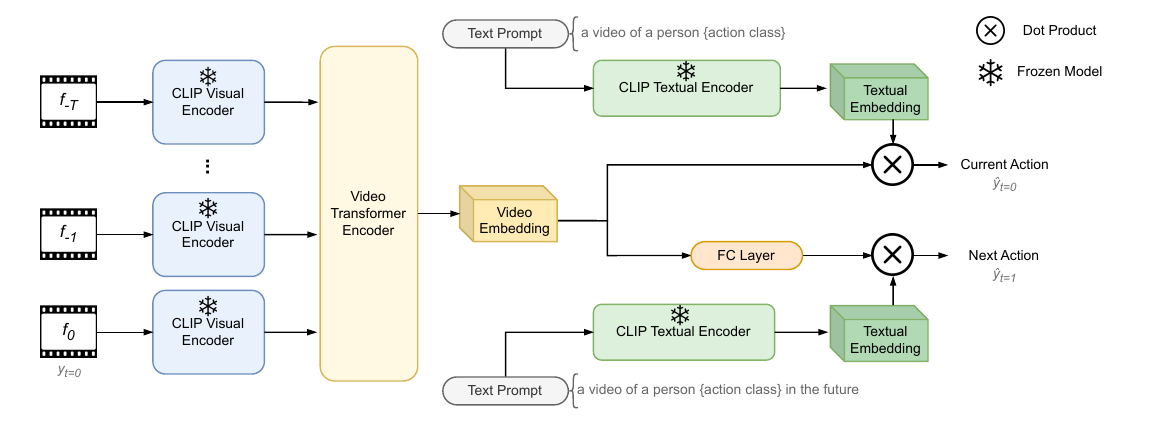}
    \caption{\textbf{Overview of TOAD.} Video features are extracted from downsampled frames (blue) and temporally aggregated using a video transformer encoder (yellow). The current action is predicted by computing the dot product between the video embedding and the textual embedding of the current action, obtained from CLIP's textual encoder (green), ensuring alignment with visual features. Since the representation of future actions differs from that of the ongoing action, a fully connected layer refines the current video representation before computing the dot product with the future action's text prompt (orange). During training, feature extractors for visual and textual features remain frozen.}
    \label{fig:toad}
\end{figure*}

This section describes TOAD, the proposed text-driven online action detection architecture depicted in Figure~\ref{fig:toad}. Our work relies on the findings of~\cite{10.1007/978-3-031-61137-7_6}, and the method introduced by Wu et al.~\cite{wu2023revisiting} to initialize a classifier with textual embeddings, following the corresponding contrastive learning approach, ActionCLIP~\cite{wang2021actionclip}, which was explored for online action detection in~\cite{10.1007/978-3-031-61137-7_6}.

\subsection{Video Encoder}
To capture temporal dynamics of video, frame aggregation is a crucial step to provide a comprehensive representation. As previously remarked, transformer architectures excel at this task. Following prior work~\cite{10.1007/978-3-031-61137-7_6,wu2023revisiting,wang2021actionclip} we adopt a standard transformer encoder, consisting of 6 attention layers, each with 12 attention heads.

From the input video, frames are uniformly sampled, and visual features are extracted using CLIP’s visual encoder~\cite{radford2021learning}. This extraction step ensures alignment between visual and textual features, which is crucial for TOAD, as discussed in this section. These extracted representations are then provided to the transformer encoder. Leveraging the self-attention mechanism, the transformer effectively captures both local frame information and long-range dependencies across the video. This mechanism also enables the model to handle varying input sizes, making it adaptable to videos of different lengths. The final video representation is obtained by averaging the frame representations, thereby summarizing the temporal information across the sequence.

\subsection{Leveraging Textual Embeddings for Classification}
Prior work on text-driven online action detection~\cite{10.1007/978-3-031-61137-7_6} relies on the \emph{``pretrain-prompt-finetune''} framework presented in~\cite{wang2021actionclip}. This framework leverages CLIP~\cite{radford2021learning} for the \emph{pretrain} step, which consists of exploiting existing pretrained models. \emph{Prompt} is defined as reformulating the downstream task to act more like the upstream pretrain task, \emph{i.e.} matching textual and video embeddings for the downstream task of action recognition. Finally, results are \emph{fine-tuned} on specific datasets.

The previous multimodal framework introduced for ActionCLIP~\cite{wang2021actionclip} aimed to face the problems of traditional action recognition methods, which address the task as a classic and standard 1-of-N majority vote problem, mapping labels into numbers. Instead, by leveraging this framework, it allows modeling the task as a vision-text multimodal learning problem, enhancing the representation through the use of natural language and enabling zero-shot and few-shot transfer. However, as previously remarked in \cite{10.1007/978-3-031-61137-7_6} and also detailed by Wu et al.~\cite{wu2023revisiting}, the contrastive paradigm requires substantial computational resources to maintain a large similarity matrix that effectively represents all classes. This matrix is crucial for calculating cosine similarities, which are necessary for computing the contrastive loss.

This limitation can be overcome by initializing a frozen classifier with textual embeddings extracted from CLIP's textual encoder as presented in \cite{wu2023revisiting}. In this way, the textual information is still exploited at a significantly reduced cost. The classification operation is formulated as follows, mimicking a fully connected layer:

\vspace{-0.5cm}
\begin{align}
    z &= v \cdot t^T
\end{align}

$z \in \mathbb{R}^{B \times C}$ corresponds to the output logits for $B$ samples and $C$ classes. $v \in \mathbb{R}^{B \times d}$ is the video representation, \emph{i.e.} the video transformer encoder's output with feature dimension $d$. And $t^T \in \mathbb{R}^{d \times C}$ is the transposed textual representation for each of the classes, obtained from CLIP's textual encoder.

We explore several approaches to initialize the textual representation. First, the \emph{class name} method represents the simplest approach, encoding only the class name from the provided labels. Second, the \emph{prompt} approach introduces the class name within a predefined prompt, as illustrated in Figure~\ref{fig:toad}. Finally, we consider a combination of these features, referred to as \emph{mixed}, which is calculated as the mean of the embeddings of \emph{class name} and \emph{prompt}. After extensive experimentation across different datasets, we find that the \emph{prompt} approach consistently performs best on average.

\subsection{Future Anticipation Facilitates Online Action Detection}
In our daily lives, anticipating future events helps us better understand the current context, such as predicting the next scene while watching a movie. Similarly, online action detection architectures use future information to improve the detection of current actions. To this end, we present a supervised approach that anticipates future actions using textual embeddings.

By harnessing the capabilities of language models, we generate an additional textual embedding that represents the next action, using the prompt: "a video of a person \emph{[action class]} in the future". Since a prompt specifying the future action is required, we only experiment with \emph{prompt} initialization of textual embeddings for future anticipation.

Given that the visual features of ongoing and future actions differ, the video embeddings are processed through an additional fully connected layer with ReLU activation before classification, refining the representation for future anticipation. This process yields $\hat{z} \in \mathbb{R}^{B \times C}$, which serves as the future visual representation.

\subsection{Training}
During training, both the classifiers and feature extractors are kept frozen. Meanwhile, the remaining parameters are trained using cross-entropy loss. The class probabilities are computed as follows:

\vspace{-0.5cm}
\begin{align} 
p &= e^\tau z \\
\hat{p} &= e^\tau \hat{z} 
\end{align}

Here, $\tau$ represents a logit scale derived from CLIP's parameters. In this context, $p$ denotes the probabilities for ongoing actions, and $\hat{p}$ represents the probabilities for future actions.

The loss function for current and future supervision is defined as:

\vspace{-0.5cm}
\begin{align} 
\mathcal{L} = CE(p, y) + \lambda CE(\hat{p}, \hat{y}) 
\end{align}

where $CE(\cdot, \cdot)$ denotes the cross-entropy loss, $y$ is the true label for the current action, $\hat{y}$ is the true label for the future action, and $\lambda$ is a weighting factor for the future supervision loss set to $0.5$.
\section{Experiments} \label{sec:exp}
This section describes the experimental setup on two well-known benchmarks for online action detection. The obtained results are compared with current state-of-the-art methods, and a new benchmark for zero-shot and few-shot online action detection is presented on two different initializations of the weights. Finally, the effectiveness of the proposal is demonstrated through an extensive ablation study.

\subsection{Experimental Setup}
\noindent\textbf{Datasets.} We conduct experiments on two widely used benchmarks, THUMOS14~\cite{thumos14} and TVSeries~\cite{de2016online}. The former includes sports videos from 20 action categories containing 200 videos for validation and 213 for testing. THUMOS14 challenges for online action detection contain drastic intra-category varieties, motion blur, short action instances, etc. Following previous work~\cite{gao2017red,Xu_2019_ICCV,Eun_2020_CVPR,Wang_2021_ICCV,xu2021long,yang2022colar,10.1145/3582649.3582656,10.1007/978-3-031-61137-7_6}, we train the model on the validation set and evaluate performance on the test set.

TVSeries gathers about 16 hours of video from 6 popular TV series. In total there are 6231 instances belonging to 30 different classes comprising daily actions. This benchmark exhibits challenges such as the temporal overlapping of actions, unconstrained perspectives and a large number of background frames. We use the default train/test splits provided in the dataset.

\vspace{0.25cm}\noindent\textbf{Metrics.} We report per-frame mean Average Precision (mAP) to evaluate on the THUMOS14 dataset, and per-frame mean calibrated Average Precision (mcAP) to evaluate on TVSeries to evaluate TOAD's performance following~\cite{Wang_2021_ICCV,xu2021long,zhao2022testra}. 

mAP is a widely used metric that needs the Average Precision (AP) of each class. The calibrated Average Precision was introduced in \cite{de2016online} to compensate for the ratio of positive versus negative frames. In other words, to reduce the probability of falsely detecting background frames with higher confidence than true positives. It is calculated as follows:

\begin{equation}
    cPrec = \frac{TP}{TP + \frac{FP}{w}} = \frac{w \cdot TP}{w \cdot TP + FP}
\end{equation}

where $w$ is the ratio between negative and positive frames.

\vspace{0.25cm}\noindent\textbf{Implementation Details.} We experiment with different visual encoders from CLIP~\cite{radford2021learning}, with ViT large with patch size 14 (ViT-L/14), and its corresponding textual encoder yielding the best results. Frames are resized to $224 \times 224$, and a downsampling rate of 6 is used before uniform sampling. We experiment with different segment lengths as input (8, 16, 32, 64). The method has been implemented in PyTorch and all experiments are performed on Nvidia RTX 4090 GPUs. During training AdamW optimizer with weight decay $0.2$ is used with a base learning rate of  $5 \times 10^{-5}$. The learning rate is warmed up during $5$ epochs and follows a cosine schedule afterwards. Models are trained for $30$ epochs with a batch size of $32$.

\subsection{Comparison with State-of-the-Art}
Table~\ref{tab:compare} presents a comparison of TOAD with other state-of-the-art methods for online action detection. Our approach shows superior performance on the THUMOS14 dataset compared to previous methods. However, it shows relatively inferior performance on the TVSeries dataset. This discrepancy is due to our reliance on RGB-only inputs, while previous work effectively uses optical flow, which has been shown to significantly improve performance, as highlighted in~\cite{de2016online}. In this study, motion vectors outperformed both convolutional and recurrent neural networks that relied on RGB-only inputs.

These results are based on the optimal settings identified in the ablation study, which will be detailed in the following sections. For THUMOS14, we utilize 64 input frames with the \emph{prompt} classifier, while for TVSeries, we use 8 input frames with the \emph{mixed} classifier.

\begin{table}[htbp]
\centering
\caption{Comparison with state-of-the-art methods. The dashed line separates VLMs from standard classifiers. Bold indicates global best, underline indicates best VLM.}
\begin{tabular}{lcc}
\toprule
Method & \begin{tabular}[c]{@{}c@{}}THUMOS14\\ (\% mAP)\end{tabular} & \begin{tabular}[c]{@{}c@{}}TVSeries\\ (\% mcAP)\end{tabular} \\ \midrule
RED \cite{gao2017red}                        & 45.3           & 79.2            \\
TRN \cite{Xu_2019_ICCV}                   & 47.2           & 83.7            \\
IDN \cite{Eun_2020_CVPR}                & 60.3           & 86.1            \\
OadTR \cite{Wang_2021_ICCV}                & 65.2           & 87.2            \\
LSTR \cite{xu2021long}                & 69.5           & \textbf{89.1}            \\
TesTra \cite{zhao2022testra}                & 71.2           & -               \\
Colar \cite{yang2022colar} & 66.9           & 88.1            \\
LightTR \cite{10.1145/3582649.3582656}                    & 68.5           & 86.6 \\\hdashline
Online ActionCLIP \cite{10.1007/978-3-031-61137-7_6} & 46.7 &  75.5 \\
\textbf{TOAD} & \underline{\textbf{82.5}} & \underline{81.0} \\\bottomrule
\end{tabular}
\label{tab:compare}
\end{table}
Our results are further limited by the lack of a pre-trained model capable of effectively integrating optical flow with textual representations, as well as a limitation on the computational resources required to train such a model, as discussed in \cite{10.1007/978-3-031-61137-7_6,wu2023revisiting}.

\subsection{Zero-shot \& Few-shot Online Action Detection}
Given the zero-shot and few-shot learning capabilities of VLMs, this section presents the results obtained by leveraging two distinct models. First, we initialize the weights with the best performing THUMOS14 model to evaluate the knowledge transfer of the future anticipation mechanism. Second, due to the relatively small size of THUMOS14 dataset which may limit efficient knowledge transfer, we exploit a pre-trained model on Kinetics-400 dataset~\cite{kay2017kinetics} from \cite{wu2023revisiting}.

We maintain the experimental settings that produced the best results as in the previous section, adjusting only the number of training samples for few-shot learning experiments. Specifically, we conduct experiments with 1, 2, 4, and 8 training samples. Following the approach in~\cite{wu2023revisiting}, we repeat the few-shot samples to ensure an equal number of iterations as their full-shot counterparts, providing a fair comparison.

\vspace{0.25cm}\noindent\textbf{THUMOS14 Initialization:}
Table~\ref{tab:fs-thumos} contains the results on the TVSeries dataset (8 input frames with \emph{mixed} classifier) initialized to the best THUMOS14 model (64 inputs frames with \emph{prompt} classifier). Despite the reduced number of samples per class, the model maintains strong performance and effectively transfers the knowledge acquired for anticipating future actions. Moreover, it is able to effectively handle a different initialization of the classifier and number of input frames thanks to the ability of the video transformer encoder.

\begin{table}[htb]
\centering
\caption{Zero-shot and Few-shot results on TVSeries (\% mcAP) when initialized to the best THUMOS14 model.}
\begin{tabular}{cccccc}
\toprule
\multirow{2}{*}{Future} & \multirow{2}{*}{Zero-Shot} & \multicolumn{4}{c}{Few-Shot} \\\cline{3-6}
& & \rule{0pt}{2.5ex} 1 & 2 & 4 & 8 \\\midrule
- & 50.97 & \textbf{58.19} & 59.51 & \textbf{63.52} & 66.17 \\
\checkmark & \textbf{50.98} & 52.48  & \textbf{62.00}  & 60.67 & \textbf{66.31} \\\bottomrule
\end{tabular}
\label{tab:fs-thumos}
\end{table}

\vspace{0.25cm}\noindent\textbf{Kinetics-400 initialization:}
The performance in Table~\ref{tab:fs-thumos} is limited due to the size of THUMOS14 dataset. For this reason, and to provide a stronger baseline for zero-shot and few-shot online action detection, we initialize the transformer encoder layers with a pre-trained model on Kinetics-400 dataset~\cite{kay2017kinetics} from \cite{wu2023revisiting}. Note that these experiments do not include the future anticipation mechanism, as the Kinetics-400 dataset is annotated for action recognition. Results are presented in Table~\ref{tab:fs-k400}. This model maintains zero-shot and few-shot capabilities on the THUMOS14 dataset, though these are limited on TVSeries. Despite the improvement compared to THUMOS14 initialization, there is a small improvement between zero-shot and few-shot with 8 training samples. Training with 1 or 2 samples per class has a negative impact on the results due to the large variability inherent in this benchmark.

\begin{table}[htbp]
\centering
\caption{Zero-Shot and Few-Shot results on THUMOS14 and TVSeries when initialized to Kinetics-400 weights.}
\begin{tabular}{ccc}
\toprule
\# Training Samples & \begin{tabular}[c]{@{}c@{}}THUMOS14\\ (\% mAP)\end{tabular} & \begin{tabular}[c]{@{}c@{}}TVSeries\\ (\% mcAP)\end{tabular} \\ \midrule
Zero-Shot & 59.36 & 71.29 \\
Few-Shot 1 & 62.96 & 68.85 \\
Few-Shot 2 & 65.86 & 70.67 \\
Few-Shot 4 & 70.95 & 72.3 \\
Few-Shot 8 & 72.25 & 72.44 \\\bottomrule
\end{tabular}
\label{tab:fs-k400}
\end{table}

\subsection{Ablation Study}
To determine the optimal settings for our model and assess the effectiveness of our proposal, we conduct extensive ablation experiments on each component of the proposed architecture. Unless specified otherwise, all experiments are performed with an input length of 64 for THUMOS14, and 8 for TVSeries, \emph{prompt} initialization for the classifier, including future information, and ViT-L/14 as RGB feature extractor.

\vspace{0.25cm}\noindent\textbf{What is the best RGB encoder?}
Table~\ref{tab:rgb} presents the results on the performance of different RGB encoders within CLIP's architecture to determine the most effective model for online action detection. From the available pre-trained models in CLIP, we compare ResNet101~\cite{he2016deep} and three variants of the Vision Transformer (ViT)~\cite{dosovitskiy2021image}, specifically ViT-B/32, ViT-B/16, and ViT-L/14. Each encoder extracts features from downsampled frames, which are subsequently aggregated to generate a representation suitable for action detection. The results indicate a clear performance increase as we move from ResNet101 to larger ViT models, highlighting the importance of model capacity and attention-based architectures in capturing the temporal and spatial information critical for accurate online action detection. It is also worth mentioning that the performance of TOAD will likely improve by unfreezing these layers and fine-tuning them end-to-end.

\begin{table}[htbp]
\centering
\caption{Ablation on CLIP's RGB encoder.}
\begin{tabular}{ccc}
\toprule
CLIP Encoder & \begin{tabular}[c]{@{}c@{}}THUMOS14\\ (\% mAP)\end{tabular} & \begin{tabular}[c]{@{}c@{}}TVSeries\\ (\% mcAP)\end{tabular} \\ \midrule
ResNet101 & 66.64 & 71.97 \\
ViT-B/32 & 75.18 & 74.85 \\
ViT-B/16 & 78.80 & 76.13 \\
ViT-L/14 & \textbf{82.46} & \textbf{80.63} \\\bottomrule
\end{tabular}
\label{tab:rgb}
\end{table}

\vspace{0.25cm}\noindent\textbf{What is the best input length?}
The input length is dependent on each dataset and the duration of the annotated actions. In this work, as previously stated, we experiment with 8, 16, 32 and 64 uniformly sampled frames, after a downsampling with a rate of 6. This corresponds to visualizing between 2 and 16 seconds of video at 30 frames per second.

The results, presented in Table~\ref{tab:len}, indicate that for THUMOS14 dataset, using a higher number of input frames generally leads to improved model performance. However, on TVSeries benchmark, the trend is reversed, with fewer input frames yielding the best results. This is attributed to a shorter duration of the annotated actions on the TVSeries dataset.

\begin{table}[htbp]
\centering
\caption{Ablation on the number of input frames.}
\begin{tabular}{ccc}
\toprule
\# Frames & \begin{tabular}[c]{@{}c@{}}THUMOS14\\ (\% mAP)\end{tabular} & \begin{tabular}[c]{@{}c@{}}TVSeries\\ (\% mcAP)\end{tabular} \\ \midrule
8 & 76.94 & \textbf{80.63} \\
16 & 79.02 & 79.34 \\
32 & 76.88 & 78.03 \\
64 & \textbf{82.46} & 78.97 \\\bottomrule
\end{tabular}
\label{tab:len}
\end{table}

\vspace{0.25cm}\noindent\textbf{What is the best classifier initialization?}
Table~\ref{tab:fut} compares different strategies to generate the textual embeddings used for classification. Results show that \emph{prompt} provides significantly better results on the THUMOS14 dataset in every scenario. Conversely, results on the TVSeries benchmark, despite less consistent when using future information, show minor differences between the different strategies.

\vspace{0.25cm}\noindent\textbf{What is the effect of future information?}
In this work a novel mechanism for the anticipation of the next action has been implemented leveraging the capabilities of VLMs. As shown in Table~\ref{tab:fut}, the performance of the classifiers significantly improves with this mechanism, demonstrating the efficacy of the presented method. The results indicate that the integration of future predictions enables classifiers to make more accurate predictions of the ongoing actions.

Depending on the prompt used to extract text embeddings for initializing the classification layers, we find that encoding only the \emph{Class Name} yields the lowest performance. In contrast, incorporating the class name within a \emph{Prompt} significantly improves results, achieving the best performance on average. A mean of the \emph{Prompt} and \emph{Class Name} initializations, referred to as \emph{Mixed}, yields intermediate results on the THUMOS14 dataset. For the TVSeries dataset, however, the \emph{Mixed} classifier achieves the highest performance, though the differences between methods are marginal.

\begin{table}[htbp]
\centering
\caption{Ablation on the effect of future anticipation and the initialization of the classifier.}
\begin{tabular}{cccc}
\toprule
Future & Classifier & \begin{tabular}[c]{@{}c@{}}THUMOS14\\ (\% mAP)\end{tabular} & \begin{tabular}[c]{@{}c@{}}TVSeries\\ (\% mcAP)\end{tabular} \\ \midrule
\multirow{3}{*}{-} & \emph{Prompt} & \underline{78.55} & 79.4 \\
& \emph{Class Name} & 59.04 & \underline{79.8} \\
& \emph{Mixed} & 67.39 & 79.7 \\\hdashline
\multirow{3}{*}{\checkmark} & \emph{Prompt} & \textbf{82.46} & 80.63 \\
& \emph{Class Name} & 68.24 & 80.37 \\
& \emph{Mixed} & 81.95 & \textbf{81.00} \\\bottomrule
\end{tabular}
\label{tab:fut}
\end{table}

\section{Conclusion} \label{sec:concl}
In this work, we introduced TOAD, the first text-driven model for online action detection. Building on the promising results from previous studies \cite{10.1007/978-3-031-61137-7_6,wu2023revisiting}, we leverage CLIP's textual embeddings for the classification of aggregated video frames from CLIP's visual encoder. This approach harnesses the powerful capabilities of vision-language models (VLMs) without relying on computationally expensive contrastive learning techniques. By predicting actions further into the future, TOAD has demonstrated improved performance in online action detection while maintaining robust zero-shot and few-shot learning capabilities.

TOAD outperforms previous methods on the THUMOS14 dataset. It also sets a new baseline for zero-shot and few-shot online action detection on THUMOS14 and TVSeries, paving the way for future research in this field. Nonetheless, given that its performance is diminished on regular online action detection on TVSeries, it is important to acknowledge certain limitations that may be addressed in future work. Specifically, (1) TOAD exhibits lower performance on the TVSeries dataset, likely due to the absence of optical flow data, which plays a critical role in capturing fine-grained motion information in this dataset, as seen in the baseline results presented in \cite{de2016online} where using motion vectors outperforms CNN- and LSTM-based architectures. Additionally, (2) the hardware requirements for end-to-end training, as seen in \cite{wu2023revisiting,10.1007/978-3-031-61137-7_6}, limit the performance of our approach. This is because TOAD relies on frozen features rather than fine-tuning parameters, which could otherwise enhance the model's adaptability and accuracy. Finally, by further exploiting the capabilities of VLMs, future research could also focus on enhancing the explainability \cite{gorriz2023computational} of the model's decisions, thereby ensuring transparency and interpretability in its functioning.

\section*{Acknowledgments}
We would like to thank CIAICO/2022/132 Consolidated group project ``AI4-Health'' funded by the Valencian government, and International Center for Aging Research ICAR funded project ``IASISTEM''. This work has also been supported by a Spanish national and a regional grants for PhD studies, FPU21/00414, CIACIF/2021/430 and CIACIF/2022/175.

\bibliographystyle{IEEEtran}
\bibliography{biblio}

\begin{thebibliography}{10}
\providecommand{\url}[1]{#1}
\csname url@samestyle\endcsname
\providecommand{\newblock}{\relax}
\providecommand{\bibinfo}[2]{#2}
\providecommand{\BIBentrySTDinterwordspacing}{\spaceskip=0pt\relax}
\providecommand{\BIBentryALTinterwordstretchfactor}{4}
\providecommand{\BIBentryALTinterwordspacing}{\spaceskip=\fontdimen2\font plus
\BIBentryALTinterwordstretchfactor\fontdimen3\font minus \fontdimen4\font\relax}
\providecommand{\BIBforeignlanguage}[2]{{%
\expandafter\ifx\csname l@#1\endcsname\relax
\typeout{** WARNING: IEEEtran.bst: No hyphenation pattern has been}%
\typeout{** loaded for the language `#1'. Using the pattern for}%
\typeout{** the default language instead.}%
\else
\language=\csname l@#1\endcsname
\fi
#2}}
\providecommand{\BIBdecl}{\relax}
\BIBdecl

\bibitem{oad-surveillance}
P.~Ni, S.~Lv, X.~Zhu, Q.~Cao, and W.~Zhang, ``A light-weight on-line action detection with hand trajectories for industrial surveillance,'' \emph{Digital Communications and Networks}, vol.~7, no.~1, pp. 157--166, 2021.

\bibitem{GARCIARODRIGUEZ20114413}
J.~García-Rodríguez and J.~M. García-Chamizo, ``Surveillance and human–computer interaction applications of self-growing models,'' \emph{Applied Soft Computing}, vol.~11, no.~7, pp. 4413--4431, 2011.

\bibitem{Kim_2019_CVPR}
J.~Kim, T.~Misu, Y.-T. Chen, A.~Tawari, and J.~Canny, ``Grounding human-to-vehicle advice for self-driving vehicles,'' in \emph{CVPR}, June 2019.

\bibitem{ramanishka2018CVPR}
V.~Ramanishka, Y.-T. Chen \emph{et~al.}, ``Toward driving scene understanding: A dataset for learning driver behavior and causal reasoning,'' in \emph{CVPR}, 2018.

\bibitem{oad-robot}
L.~Tong, H.~Ma, Q.~Lin, J.~He, and L.~Peng, ``A novel deep learning bi-gru-i model for real-time human activity recognition using inertial sensors,'' \emph{IEEE Sensors Journal}, vol.~22, no.~6, pp. 6164--6174, 2022.

\bibitem{GOMEZDONOSO2017105}
F.~Gomez-Donoso, S.~Orts-Escolano, A.~Garcia-Garcia, J.~Garcia-Rodriguez, J.~A. Castro-Vargas, S.~Ovidiu-Oprea, and M.~Cazorla, ``A robotic platform for customized and interactive rehabilitation of persons with disabilities,'' \emph{Pattern Recognition Letters}, vol.~99, pp. 105--113, 2017.

\bibitem{de2016online}
R.~De~Geest, E.~Gavves, A.~Ghodrati, Z.~Li, C.~Snoek, and T.~Tuytelaars, ``Online action detection,'' in \emph{Computer Vision--ECCV 2016: 14th European Conference, Amsterdam, The Netherlands, October 11-14, 2016, Proceedings, Part V 14}.\hskip 1em plus 0.5em minus 0.4em\relax Springer, 2016, pp. 269--284.

\bibitem{Eun_2020_CVPR}
H.~Eun, J.~Moon, J.~Park, C.~Jung, and C.~Kim, ``Learning to discriminate information for online action detection,'' in \emph{CVPR}, 2020.

\bibitem{gao2017red}
J.~Gao, Z.~Yang, and R.~Nevatia, ``Red: Reinforced encoder-decoder networks for action anticipation,'' \emph{arXiv:1707.04818}, 2017.

\bibitem{Xu_2019_ICCV}
M.~Xu, M.~Gao, Y.-T. Chen, L.~S. Davis, and D.~J. Crandall, ``Temporal recurrent networks for online action detection,'' in \emph{ICCV}, 2019.

\bibitem{An_2023_ICCV}
J.~An, H.~Kang, S.~H. Han, M.-H. Yang, and S.~J. Kim, ``Miniroad: Minimal rnn framework for online action detection,'' in \emph{ICCV}, October 2023, pp. 10\,341--10\,350.

\bibitem{zhao2023survey}
W.~X. Zhao \emph{et~al.}, ``A survey of large language models,'' \emph{arXiv:2303.18223}, 2023.

\bibitem{dosovitskiy2021image}
A.~Dosovitskiy, L.~Beyer, A.~Kolesnikov, D.~Weissenborn, X.~Zhai, T.~Unterthiner, M.~Dehghani, M.~Minderer, G.~Heigold, S.~Gelly, J.~Uszkoreit, and N.~Houlsby, ``An image is worth 16x16 words: Transformers for image recognition at scale,'' \emph{arXiv:2010.11929}, 2021.

\bibitem{arnab2021vivit}
A.~Arnab, M.~Dehghani, G.~Heigold, C.~Sun, M.~Lučić, and C.~Schmid, ``Vivit: A video vision transformer,'' \emph{arXiv:2103.15691}, 2021.

\bibitem{piergiovanni2022rethinking}
A.~Piergiovanni, W.~Kuo, and A.~Angelova, ``Rethinking video vits: Sparse video tubes for joint image and video learning,'' \emph{arXiv:2212.03229}, 2022.

\bibitem{Wang_2021_ICCV}
X.~Wang, S.~Zhang, Z.~Qing, Y.~Shao, Z.~Zuo, C.~Gao, and N.~Sang, ``Oadtr: Online action detection with transformers,'' in \emph{ICCV}, October 2021, pp. 7565--7575.

\bibitem{10.1145/3582649.3582656}
R.~Li, L.~Yan, Y.~Peng, and L.~Qing, ``Lighter transformer for online action detection,'' in \emph{Proceedings of the 2023 6th International Conference on Image and Graphics Processing}, ser. ICIGP 2023.\hskip 1em plus 0.5em minus 0.4em\relax ACM, Jan. 2023, p. 161–167.

\bibitem{xu2021long}
M.~Xu, Y.~Xiong, H.~Chen, X.~Li, W.~Xia, Z.~Tu, and S.~Soatto, ``Long short-term transformer for online action detection,'' in \emph{NeurIPS}, 2021.

\bibitem{zhao2022testra}
Y.~Zhao and P.~Kr{\"a}henb{\"u}hl, ``Real-time online video detection with temporal smoothing transformers,'' in \emph{European Conference on Computer Vision (ECCV)}, 2022.

\bibitem{wang2021actionclip}
M.~Wang, J.~Xing, and Y.~Liu, ``Actionclip: A new paradigm for video action recognition,'' \emph{arXiv:2109.08472}, 2021.

\bibitem{bike}
W.~Wu, X.~Wang, H.~Luo, J.~Wang, Y.~Yang, and W.~Ouyang, ``Bidirectional cross-modal knowledge exploration for video recognition with pre-trained vision-language models,'' in \emph{Proceedings of the IEEE/CVF Conference on Computer Vision and Pattern Recognition}, 2023.

\bibitem{radford2021learning}
A.~Radford, J.~W. Kim, C.~Hallacy, A.~Ramesh, G.~Goh, S.~Agarwal, G.~Sastry, A.~Askell, P.~Mishkin, J.~Clark, G.~Krueger, and I.~Sutskever, ``Learning transferable visual models from natural language supervision,'' \emph{arXiv:2103.00020}, 2021.

\bibitem{cheng2023vindlu}
F.~Cheng, X.~Wang, J.~Lei, D.~Crandall, M.~Bansal, and G.~Bertasius, ``Vindlu: A recipe for effective video-and-language pretraining,'' \emph{arXiv:2212.05051}, 2023.

\bibitem{ju2022prompting}
C.~Ju, T.~Han, K.~Zheng, Y.~Zhang, and W.~Xie, ``Prompting visual-language models for efficient video understanding,'' \emph{arXiv:2112.04478}, 2022.

\bibitem{papalampidi2023simple}
P.~Papalampidi, S.~Koppula, S.~Pathak, J.~Chiu, J.~Heyward, V.~Patraucean, J.~Shen, A.~Miech, A.~Zisserman, and A.~Nematzdeh, ``A simple recipe for contrastively pre-training video-first encoders beyond 16 frames,'' \emph{arXiv:2312.07395}, 2023.

\bibitem{li2023strong}
Z.~Li \emph{et~al.}, ``A strong baseline for temporal video-text alignment,'' \emph{arXiv:2312.14055}, 2023.

\bibitem{wu2023revisiting}
W.~Wu, Z.~Sun, and W.~Ouyang, ``Revisiting classifier: Transferring vision-language models for video recognition,'' in \emph{AAAI Conf.}, vol.~37, 2023, pp. 2847--2855.

\bibitem{10.1007/978-3-031-61137-7_6}
M.~Benavent-Lledo, D.~Mulero-P{\'e}rez, D.~Ortiz-Perez, J.~Garcia-Rodriguez, and S.~Orts-Escolano, ``Exploring text-driven approaches for online action detection,'' in \emph{Bioinspired Systems for Translational Applications: From Robotics to Social Engineering}, J.~M. Ferr{\'a}ndez~Vicente, M.~Val~Calvo, and H.~Adeli, Eds.\hskip 1em plus 0.5em minus 0.4em\relax Cham: Springer Nature Switzerland, 2024, pp. 55--64.

\bibitem{brown2020language}
T.~B. Brown, B.~Mann, N.~Ryder \emph{et~al.}, ``Language models are few-shot learners,'' \emph{arXiv:2005.14165}, 2020.

\bibitem{touvron2023llama}
H.~Touvron, L.~Martin, K.~Stone \emph{et~al.}, ``Llama 2: Open foundation and fine-tuned chat models,'' \emph{arXiv:2307.09288}, 2023.

\bibitem{devlin2018bert}
J.~Devlin, M.-W. Chang, K.~Lee, and K.~Toutanova, ``Bert: Pre-training of deep bidirectional transformers for language understanding,'' \emph{arXiv:1810.04805}, 2018.

\bibitem{8627985}
S.~Li, Z.~Tao, K.~Li, and Y.~Fu, ``Visual to text: Survey of image and video captioning,'' \emph{IEEE Transactions on Emerging Topics in Computational Intelligence}, vol.~3, no.~4, pp. 297--312, 2019.

\bibitem{xu2021videoclip}
H.~Xu, G.~Ghosh, P.-Y. Huang, D.~Okhonko, A.~Aghajanyan, F.~Metze, L.~Zettlemoyer, and C.~Feichtenhofer, ``Videoclip: Contrastive pre-training for zero-shot video-text understanding,'' \emph{arXiv:2109.14084}, 2021.

\bibitem{Hu2022}
X.~Hu, J.~Dai, M.~Li, C.~Peng, Y.~Li, and S.~Du, ``Online human action detection and anticipation in videos: A survey,'' \emph{Neurocomputing}, vol. 491, p. 395–413, Jun. 2022.

\bibitem{yang2022colar}
L.~Yang, J.~Han, and D.~Zhang, ``Colar: Effective and efficient online action detection by consulting exemplars,'' in \emph{CVPR}, 2022.

\bibitem{Gao_2021_CVPR}
M.~Gao, Y.~Zhou, R.~Xu, R.~Socher, and C.~Xiong, ``Woad: Weakly supervised online action detection in untrimmed videos,'' in \emph{CVPR}, June 2021, pp. 1915--1923.

\bibitem{thumos14}
Y.-G. Jiang, J.~Liu \emph{et~al.}, ``Thumos challenge: Action recognition with a large number of classes,'' 2014.

\bibitem{kay2017kinetics}
W.~Kay, J.~Carreira, K.~Simonyan, B.~Zhang, C.~Hillier, S.~Vijayanarasimhan, F.~Viola, T.~Green, T.~Back, P.~Natsev \emph{et~al.}, ``The kinetics human action video dataset,'' \emph{arXiv preprint arXiv:1705.06950}, 2017.

\bibitem{he2016deep}
K.~He, X.~Zhang, S.~Ren, and J.~Sun, ``Deep residual learning for image recognition,'' in \emph{Proceedings of the IEEE conference on computer vision and pattern recognition}, 2016, pp. 770--778.

\bibitem{gorriz2023computational}
J.~M. G{\'o}rriz, I.~{\'A}lvarez-Ill{\'a}n, A.~{\'A}lvarez-Marquina, J.~E. Arco, M.~Atzmueller, F.~Ballarini, E.~Barakova, G.~Bologna, P.~Bonomini, G.~Castellanos-Dominguez \emph{et~al.}, ``Computational approaches to explainable artificial intelligence: advances in theory, applications and trends,'' \emph{Information Fusion}, vol. 100, p. 101945, 2023.

\end{thebibliography}

\end{document}